\documentclass[letterpaper, 10 pt, conference]{ieeeconf}  

\IEEEoverridecommandlockouts                              
                                                          
\usepackage{moreverb,url}
\usepackage{graphicx}

\usepackage[colorlinks,bookmarksopen,bookmarksnumbered,citecolor=red,urlcolor=red]{hyperref}
\usepackage{booktabs}
\usepackage[table]{xcolor}

\usepackage{cite}
\usepackage[cmex10]{amsmath}
\usepackage{algorithmic}
\usepackage{array}
\usepackage[tight,footnotesize]{subfigure}
\usepackage{flushend}

\begin{document}

\title{\LARGE \bf The Limits and Potentials of Deep Learning for Robotics}

\author{Niko S\"underhauf${}^1$, 
     Oliver Brock${}^2$,
     Walter Scheirer${}^3$,
     Raia Hadsell${}^4$, 
     Dieter Fox${}^5$, \\
     J\"urgen Leitner${}^1$,
     Ben Upcroft${}^6$,   
     Pieter Abbeel${}^7$,
     Wolfram Burgard${}^8$,    
     Michael Milford${}^1$,
     Peter Corke${}^1$
\thanks{${}^1$Australian Centre for Robotic Vision, Queensland University of Technology (QUT), Brisbane, Australia.}
\thanks{${}^2$Robotics and Biology Laboratory, Technische Universit\"at Berlin, Germany.}
\thanks{${}^3$Department of Computer Science and Engineering, University of Notre Dame, IN, USA.}
\thanks{${}^4$DeepMind, London, U.K.}
\thanks{${}^5$Paul G.~Allen School of Computer Science \& Engineering, University of Washington, WA, USA.}
\thanks{${}^6$Oxbotica Ltd., Oxford, U.K.}
\thanks{${}^7$UC Berkeley, Department of Electrical Engineering and Computer Sciences, CA, USA.}
\thanks{${}^8$Wolfram is with the Department of Computer Science, University of Freiburg, Germany.}
\thanks{Corresponding author: {\tt\small niko.suenderhauf@qut.edu.au}}
}

\maketitle

\begin{abstract}
The application of deep learning in robotics leads to very specific problems and research questions that are typically not addressed by the computer vision and machine learning communities. In this paper we discuss a number of robotics-specific learning, reasoning, and embodiment challenges for deep learning. We explain the need for better evaluation metrics, highlight the importance and unique challenges for deep robotic learning in simulation, and explore the spectrum between purely data-driven and model-driven approaches. We hope this paper provides a motivating overview of important research directions to overcome the current limitations, and help fulfill the promising potentials of deep learning in robotics.
\end{abstract}

\section{Introduction}

A robot is an inherently \emph{active} agent that interacts with the real world, and often operates in uncontrolled or detrimental conditions. Robots have to perceive, decide, plan, and execute actions -- all based on incomplete and uncertain knowledge. Mistakes can lead to potentially catastrophic results that will not only endanger the success of the robot's mission, but can even put human lives at stake, e.g. if the robot is a driverless car.

The application of deep learning in robotics therefore motivates research questions that differ from those typically addressed in computer vision: How much trust can we put in the predictions of a deep learning system when misclassifications can have catastrophic consequences? How can we estimate the uncertainty in a deep network's predictions and how can we fuse these predictions with prior knowledge and other sensors in a probabilistic framework? How well does deep learning perform in realistic unconstrained open-set scenarios where objects of unknown class and appearance are regularly encountered? 

If we want to use data-driven learning approaches to generate
motor commands for robots to move and act in the world, 
we are faced with additional challenging questions:
How can we generate enough high-quality training data? Do we rely on data solely collected on robots in real-world scenarios or do we require data augmentation through simulation? How can we ensure the learned policies transfer well to different situations, from simulation to reality, or between different robots?

This leads to further fundamental questions: How can the structure, the constraints, and the physical laws that govern robotic tasks in the real world be leveraged and exploited by a deep learning system? Is there a fundamental difference between model-driven and data-driven problem solving, or are these rather two ends of a spectrum?  

This paper explores some of the challenges, limits, and potentials for deep learning in robotics. The invited speakers and organizers of the workshop on \emph{The Limits and Potentials of Deep Learning for Robotics} at the 2016 edition of the \emph{Robotics: Science and Systems} (RSS) conference~\cite{rss16-ws-skeptics} provide their thoughts and opinions, and point out open research problems and questions that are yet to be answered. We hope this paper will offer the interested reader an overview of where we believe important research needs to be done, and where deep learning can have an even bigger impact in robotics over the coming years.

\section{Challenges for Deep Learning in Robotic Vision}

A robot is an inherently \emph{active} agent that acts \emph{in}, and interacts \emph{with} the physical real world. 
It perceives the world with its different sensors, builds a coherent model of the world and updates this model over time, but ultimately a robot has to make decisions, plan actions, and execute these actions to fulfill a useful task.

This is where \emph{robotic vision} differs from computer vision. For robotic vision, perception is only one part of a more complex, embodied, active, and goal-driven system.  \emph{Robotic} vision therefore has to take into account that its \emph{immediate} outputs (object detection, segmentation, depth estimates, 3D reconstruction, a description of the scene, and so on), will ultimately result in \emph{actions} in the real world. In a simplified view, while computer vision takes images and translates them into information, robotic vision translates images into actions.

This fundamental difference between robotic vision and computer vision motivates a number of research challenges along three conceptually orthogonal axes: \emph{learning}, \emph{embodiment}, and \emph{reasoning}. We position individual challenges along these axes according to their increasing complexity, and their dependencies. 
Tables \ref{tab:challenges1}--\ref{tab:challenges3} summarize the challenges.

\begin{figure}[tb]
    \centering
    \includegraphics[width=0.8\linewidth]{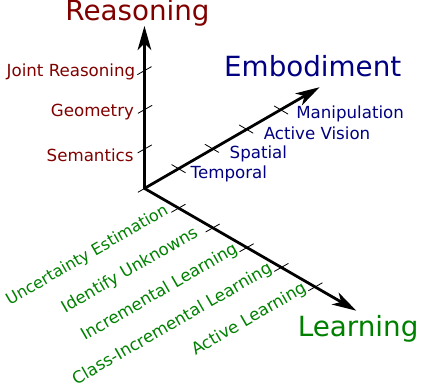}
    \caption{Current Challenges for Deep Learning in Robotic Vision. We can categorize these challenges into three conceptually orthogonal axes: learning, embodiment, and reasoning.}
    \label{fig:challenges}
\end{figure}

\subsection{Learning Challenges}

Along this axis we position challenges that are specific for (deep) machine learning in a robotic vision context. These challenges comprise problems arising from deployment in open-set conditions, two flavours of incremental learning, and active learning.

\subsubsection{Uncertainty Estimation}
\label{sec:uncertainty}

In order to fully integrate deep learning into robotics, it is important that deep learning systems can reliably estimate the uncertainty in their predictions. This would allow robots to treat a deep neural network like any other sensor, and use the established Bayesian techniques \cite{Thrun05, Kummerle11, Kaess12} to fuse the network's predictions with prior knowledge or other sensor measurements, or to accumulate information over time. 
Deep learning systems, e.g. for classification or detection, typically return scores from their softmax layers that are proportional to the system's confidence, but are \emph{not} calibrated probabilities, and therefore not useable in a Bayesian sensor fusion framework.

Current approaches towards uncertainty estimation for deep learning are calibration techniques \cite{hendrycks2017baseline,guo2017calibration}, or Bayesian deep learning \cite{mackay1992practical, neal1995bayesian} with approximations such as Dropout Sampling~\cite{gal2016dropout, kendall2017uncertainties} or ensemble methods \cite{lakshminarayanan2017simple}.

\begin{table*}[t]
\centering
\caption{Learning Challenges for Robotic Vision}
\label{tab:challenges1}
\begin{tabular}{@{}p{0.03\linewidth}p{0.15\linewidth}p{0.75\linewidth}@{}}
 \toprule 
 Level & Name & Description \\ \midrule
 5 & Active Learning & The system is able to select the most informative samples for incremental learning on its own in a data-efficient way, e.g. by utilizing its estimated uncertainty in a prediction. It can ask the user to provide labels.\\
 4 & Class-Incremental Learning & The system can learn \emph{new} classes, preferably using low-shot or one-shot learning techniques, without catstrophic forgetting. The system requires the user to provide these new training samples along with correct class labels. \\
 3 & Incremental Learning & The system can learn off new instances of known classes to address domain adaptation or label shift. It requires the user to select these new training samples. \\
 2 & Identify Unknowns & In an open-set scenario, the robot can reliably identify instances of unknown classes and is not fooled by out-of distribution data. \\
 1 & Uncertainty Estimation & The system can correctly estimate its uncertainty and returns \emph{calibrated} confidence scores that can be used as probabilities in a Bayesian data fusion framework. Current work on Bayesian Deep Learning falls into this category.\\
 0 & Closed-Set Assumptions & The system can detect and classify objects of classes known during training. It provides uncalibrated confidence scores that are proportional to the system's belief of the label probabilities. State of the art methods, such as YOLO9000, SSD, Mask R-CNN are at this level. \\ \bottomrule
\end{tabular}
\end{table*}

\subsubsection{Identify Unknowns}
A common assumption in deep learning is that trained models will be deployed under \emph{closed-set} conditions \cite{bendale2015towards,torralba2011unbiased}, i.e. the classes encountered during deployment are known and exactly the same as during training. However, robots often have to operate in ever-changing, uncontrolled real-world environments, and will inevitably encounter instances of classes, scenarios, textures, or environmental conditions that were not covered by the training data.

In these so called \emph{open-set} conditions \cite{scheirer2013toward, bendale2015towards}, it is crucial to identify the unknowns: The perception system must not assign high-confidence scores to unknown objects or falsely recognize them as one of the known classes. If for example an object detection system is fooled by data outside of its training data distribution \cite{goodfellow2014explaining, nguyen2015deep}, the consequences for a robot acting on false, but high-confidence detections can be catastrophic.
One way to handle the open-set problem and identify unknowns is to utilize the epistemic uncertainty \cite{kendall2017uncertainties, gal2016dropout} of the model predictions to reject predictions with low confidence \cite{miller2017dropout}. 

\subsubsection{Incremental Learning}
For many robotics applications the characteristics and appearance of objects can be quite different in the deployment scenario compared to the training data. To address this domain adaptation problem \cite{csurka2017domain, patel2015visual,Ganin2015domain}, a robotic vision system should be able to learn from new training samples of known classes during deployment and adopt its internal representations accordingly.

\subsubsection{Class-Incremental Learning}
When operating in open-set conditions, the deployment scenario might contain new classes of interest that were not available during training. A robot therefore needs the capability to extend its knowledge and efficiently learn new classes without forgetting the previously learned representations \cite{goodfellow2013empirical}. This class-incremental learning would preferably be data-efficient by using one-shot \cite{lake2015human, vinyals2016matching, bertinetto2016learning, rezende2016one, santoro2016meta} or low-shot \cite{hariharan2016low, wang2016learning, finn2017model} learning techniques. Semi-supervised approaches \cite{kingma2014semi, rasmus2015semi, papandreou2015weakly} that can leverage unlabeled data are of particular interest. 

Current techniques for class-incremental learning \cite{Rebuffi_2017_CVPR, mensink2012metric} still rely on supervision in the sense that the user has to specifically tell the system which samples are new data and therefore should be incorporated. The next challenge in our list, \emph{active learning}, aims to overcome this and automatically selects new training samples from the available data. 

\subsubsection{Active Learning}
A robot should be able to select the most informative samples for incremental learning techniques on its own. Since it would have to ask the human user for the true label for these selected samples, data-efficiency is key to minimize this kind of interaction with the user. Active learning \cite{cohn1996active} can also comprise retrieving annotations from other sources such as the web. 

Some current approaches \cite{gal2017deep, dayoub2017episode} leverage the uncertainty estimation techniques based on approximate Bayesian inference (see Section~\ref{sec:uncertainty}) to choose the most informative samples.

\subsection{Embodiment Challenges}

\begin{table*}[tb]
\centering
\caption{Embodiment Challenges for Robotic Vision}
\label{tab:challenges2}
\begin{tabular}{@{}p{0.03\linewidth}p{0.15\linewidth}p{0.75\linewidth}@{}}
 \toprule
 Level & Name & Description \\ \midrule
 4 & Active Manipulation & As an extension of active vision, the system can manipulate the scene to aid perception. For example it can move an occluding object to gain information about object hidden underneath.  \\
 3 & Active Vision & The system has learned to actively control the camera movements in the world, for example it can move the camera to a better viewpoint to improve its perception confidence or better deal with occlusions. \\
 2 & Spatial Embodiment & The system can exploit aspects of spatial coherency and incorporate views of objects taken from different viewpoints to improve its perception, while handling occlusions.\\
 1 & Temporal Embodiment & The system learned that it is temporally embedded and consecutive are strongly correlated. The system can accumulate evidence over time to improve its predictions. Appearance changes over time can be coped with.\\
 0 & None & The system has no understanding of any form of embodiment and treats every image as an independent from previously seen images.    \\
 \bottomrule
\end{tabular}
\end{table*}

Embodiment is a corner stone of what constitutes robotic vision, and what sets it apart from computer vision. Along this axis we describe four embodiment challenges: understanding and utilizing temporal and spatial embodiment helps to improve perception, but also enables robotic vision to perform active vision, and even targeted manipulation of the environment to further improve perception.

\subsubsection{Temporal Embodiment}
In contrast to typical recent computer vision systems that treat every image as independent, a robotic vision system perceives a \emph{stream} of consecutive and therefore strongly correlated images. While current work on action recognition, learning from demonstration, and similar directions in computer vision work on video data, (e.g. by using recurrent neural networks or by simply stacking consecutive frames in the input layers), the potential of \emph{temporal} embodiment to improve the quality of the perception process for object detection or semantic segmentation, is currently rarely utilized:
a robotic vision system that uses its temporal embodiment can for example accumulate evidence over time -- preferably using Bayesian techniques, if uncertainty estimates are available as discussed in Section \ref{sec:uncertainty} -- or exploit small viewpoint variations that occur over time in dynamic scenes. 

The new CORe50 dataset \cite{lomonaco2017core50} is one of the few available datasets that encourages researchers to exploit temporal embodiment for object recognition, but the robotic vision research community should invest more effort to fully exploit the potentials of temporal embodiment.

A challenging aspect of temporal embodiment is that the appearance of scenes changes over time. An environment can comprise dynamic objects such as cars or pedestrians moving through the field of view of a camera. An environment can also change its appearance caused by different lighting conditions (day/night), structural changes in objects (summer/winter), or differences in the presence and pose of objects (e.g. an office during and after work hours). A robotic vision system has to cope with all of those effects.

\subsubsection{Spatial Embodiment}

In robotic vision, the camera that observes the world is part of a bigger robotic system that acts and moves in the world -- the camera is \emph{spatially} embodied. As the robot moves in its environment, the camera will observe the scene from different viewpoints, which poses both challenges and opportunities to a robotic vision system: Observing an object from different viewpoints can help to disambiguate its semantic properties, improve depth perception, or segregate an object from other objects or the background in cluttered scenes. On the other hand, occlusions and the resulting sudden appearance changes complicate visual perception and require capabilities such as object unity and object permanence \cite{piaget2013construction} that are known to develop in the human visual system \cite{goldstein2016sensation}.
 
\subsubsection{Active Vision}
One of the biggest advantages robotic vision can draw from its embodiment is the potential to \emph{control} the camera, move it, and change its viewpoint in order to improve its perception or gather additional information about the scene. This is in stark contrast to most computer vision scenarios, where the camera is a passive sensor that observes the environment from where it was placed, without any means of controlling its pose.

Some work is undertaken in the area of next-best viewpoint prediction to improve object detection \cite{Wu_2015_CVPR, doumanoglou2016recovering, malmir2017deep, atanasov2014nonmyopic} or path planning for exploration on a mobile robot \cite{bircher2016receding}, but a more holistic approach to active scene understanding is still missing from current research. Such an active robotic vision system system could  control camera movements through the world to improve the system's perception confidence, resolve ambiguities, mitigate the effect of occlusions, or reflections. 

\subsubsection{Manipulation for Perception}
As an extension of active vision, a robotic system could purposefully manipulate the scene to aid its perception. For example a robot could move occluding objects to gain information about object hidden underneath. Planning such actions will require an understanding of the geometry of the scene, the capability to reason about how certain manipulation actions will change the scene, and if those changes will positively affect the perception processes.

\begin{table*}[tb]
\centering
\caption{Reasoning Challenges for Robotic Vision}
\label{tab:challenges3}
\begin{tabular}{@{}p{0.03\linewidth}p{0.15\linewidth}p{0.75\linewidth}@{}}
 \toprule
 Level & Name & Description \\ \midrule
 3 & Joint Reasoning & The system jointly reasons about semantics and geometry in a tighly coupled way, allowing semantics and geometry to co-inform each other. \\
 2 & Object and Scene Geometry & The system learned to reason about the geometry and shape of individual objects, and about the general scene geometry, such as absolute and relative object pose, support surfaces, and object continuity under occlusions and in clutter.\\
 1 & Object and Scene Semantics & The system can exploit prior semantic knowledge to improve its performance. It can utilize priors about which objects are more likely to occur together in a scene, or how objects and overall scene type are correlated.\\
 0 & None & The system does not perform any sophisticated reasoning, e.g. it treats every detected object as independent from other objects or the overall scene. Estimates of semantics and geometry are treated as independent. \\
 \bottomrule
\end{tabular}
\end{table*}

\subsection{Reasoning Challenges}
In his influential 1867 book on Physiological Optics \cite{helmholtz1867handbuch}, Hermann von Helmholtz formulated the idea that humans use unconscious \emph{reasoning}, inference or conclusion, when processing visual information. Since then, psychologists have devised various experiments to investigate these unconscious mechanisms \cite{goldstein2016sensation}, modernized Helmholtz' original ideas \cite{rock1983logic}, and reformulated them in the framework of Bayesian inference \cite{kersten2004object}.

Inspired by their biological counterparts, we formulate the following three reasoning challenges, addressing separate and joint reasoning about the semantics and geometry of a scene and the objects therein. 

\subsubsection{Reasoning About Object and Scene Semantics}

The world around us contains many semantic regularities that humans use to aid their perception~\cite{goldstein2016sensation}: objects tend to appear more often in a certain context than in other contexts (e.g. it is more likely to find a fork in a kitchen or on a dining table, but less likely to find it in a bathroom), some objects tend to appear in groups, some objects rarely appear together in a scene, and so on. Semantic regularities also comprise the absolute pose of object in a scene, or the relative pose of an object with respect to other objects.

While the importance of semantic regularities and contextual information for human perception processes is well known in psychology~\cite{oliva2007role, goldstein2016sensation}, current object detection systems \cite{redmon2016yolo9000, he2017mask, liu2016ssd} do not exploit this rich source of information. If the many semantic regularities present in the real world can be learned or otherwise made available to the vision system in the form of prior knowledge, we can expect an improved and more robust perception performance: Context can help to disambiguate or correct predictions and detections.

The work by Lin et al. \cite{lin2013holistic} is an example of a scene understanding approach that explicitly models and exploits several semantic and geometric relations between objects and the overall scene using Conditional Random Fields. A combination of place categorization and improved object detection utilizing learned scene-object priors has been demonstrated in \cite{suenderhauf2016place}. More recent work \cite{zhang2016deepcontext} devises a method to perform holistic scene understanding using a deep neural network that learns to utilize context information from training data.

\subsubsection{Reasoning About Object and Scene Geometry}
Many applications in robotics require knowledge about the \emph{geometry} of individual objects, or the scene as a whole. Estimating the depth of the scene from a single image has become a widely researched topic \cite{garg2016unsupervised,godard2017unsupervised,liu2016learning}. Similarly, there is a lot of ongoing work on estimating the 3D structure of objects from a single or multiple views without having depth information available~\cite{choy20163d, yan2016perspective, hane2017hierarchical, zhu2017rethinking}. These methods are typically evaluated on images with only one or a few prominent and clearly separated objects. However for robotic applications, cluttered scenes are very common. 

The previously discussed problems of uncertainty estimation and coping with unknown objects apply here as well: a robotic vision system that uses the inferred geometry for example to grasp objects needs the ability to express uncertainty in the inferred object shape when planning grasp points. Similarly, it should be able to exploit its embodiment to move the camera to a better viewpoint to efficiently collect new information that enables a more accurate estimate of the object geometry.

As an extension of reasoning over individual objects, inference over the geometry of the whole scene is important for robotic vision, and closely related to the problems of object-based mapping or object-based SLAM \cite{salas2013slam++,sunderhauf2016meaningful,cadena2016past,pillai2015monocular}. Exploiting semantic and prior knowledge can help a robotic vision system to better reason about the scene structure, for example the absolute and relative poses of objects, support surfaces, and object continuity despite occlusions.

\subsubsection{Joint Reasoning about Semantics and Geometry}
The ability to extract information about objects, environmental structures, their various complex relations, and the scene geometry in complex environments under realistic, open-set conditions is increasingly important for robotics. Our final reasoning challenge for a robotic vision system therefore is the ability to reason \emph{jointly} about the semantics and the geometry of a scene and the objects therein. Since semantics and geometry can co-inform each other, a tightly coupled inference approach can be advantageous over loosely coupled approaches where reasoning over semantics and geometry is performed separately.

\section{Are We Getting Evaluation Right in Deep Learning for Robotics?}

Why doesn't real-world deep learning performance match  published performance on benchmark datasets? This is a vexing question currently facing roboticists --- and the answer has to do with the nature of evaluation in computer vision. Robotics is different from much of computer vision in that a robot must interact with a dynamic environment,  not just images or videos downloaded from the Internet. Therefore a successful algorithm must generalize to numerous novel settings, which shifts the emphasis away from a singular focus on computing the best summary statistic (e.g., average accuracy, area under the curve, precision, recall) over a canned dataset. Recent catastrophic failures of autonomous vehicles relying on convolutional neural networks~\cite{NYT5} highlight this disconnect: when a summary statistic indicates that a dataset has been solved, it does not necessarily mean that the problem itself has been solved. The consequences of this observation are potentially far reaching if algorithms are deployed without a thorough understanding of their strengths and weaknesses~\cite{anthony}. 

While there are numerous flaws lurking in the shadows of deep learning benchmarks~\cite{DBLP:journals/corr/SzegedyZSBEGF13,cox2014neural,7298640,7780542}, two key aspects are worth discussing here: 1) the open set nature of decision making in visual recognition problems related to robotics, and 2) the limitations of traditional dataset evaluation in helping us understand the capabilities of an algorithm. \textit{Open Set Recognition} refers to scenarios where incomplete knowledge of the world is present at training time, and unknown classes can be submitted to an algorithm during its operation~\cite{6365193}. It is absolutely critical to ask what the dataset isn't capturing before setting a trained model loose to perform in the real world. Moreover, if a claim is made about the human-level (or, as we've been hearing lately, superhuman-level) performance of an algorithm, human behavior across varying conditions should be the frame of reference, not just a comparison of summary statistics on a dataset. This leads us to suggest \textit{Visual Psychophysics} as a sensible alternative for evaluation.  

\subsection*{The Importance of Open Set Recognition}

In an autonomous vehicle setting, one can envision an object detection model trained to recognize other cars,  while rejecting trees, signs, telephone poles and any other non-car object in the scene. The challenge in obtaining good performance from this model is in the necessary generalization to all non-car objects --- both known and unknown. Instead of casting such a detection task as a binary decision problem like most popular classification strategies would do, it is perhaps more useful to think about it within the context of the following taxonomy~\cite{6809169}, inspired by some memorable words spoken by Donald Rumsfeld~\cite{rumsfeld-known-knowns}:
\begin{itemize}
\item \textit{Known Classes}: the classes with distinctly labeled positive training examples (also serving as negative examples for other known classes).
\item \textit{Known Unknown Classes}: labeled negative examples, not necessarily grouped into meaningful categories.
\item \textit{Unknown Unknown Classes}: classes unseen in training. These samples are the most problematic for machine learning.
\end{itemize}

Should not the feature space produced by a deep learning method help us out with the unknown classes? After all, the advantage of deep learning is the ability to learn separable feature representations that are strongly invariant to changing scene conditions. The trouble we find is not necessarily with the features themselves, but in the read-out layer used for decision making. Consider the following problems with three popular classifiers used as read-out layers for convolutional neural networks when applied to recognition tasks where unknown classes are present. A linear SVM separates the positive and negative classes by a single linear decision boundary, establishing two half-spaces. These half-spaces are infinite in extent, meaning unknown samples far from the support of known training data can receive a positive label~\cite{6809169}. The Softmax function is a common choice for multi-class classification, but computing it requires calculating a summation over all of the classes. This is not possible when unknown classes are expected at testing time~\cite{7780542}. Along these same lines, when used to make a decision, cosine similarity requires a threshold, which can only be estimated over known data. The difficulty of establishing decision boundaries that capture a large measure of intraclass variance while rejecting unknown classes underpins several well-known deficiencies in deep learning architectures~\cite{DBLP:journals/corr/SzegedyZSBEGF13,7298640}. 

It is readily apparent that we do not understand decision boundary modeling as well as we should. Accordingly, we suggest that researchers give more attention to decision making at an algorithmic level to address the limitations of existing classification mechanisms. What is needed is a new class of machine learning algorithms that minimize the risk of the unknown. Preliminary work exploring this idea has included slab-based linear classifiers to limit the risk of half-spaces~\cite{6365193}, nearest non-outlier models~\cite{bendale2015towards}, and extreme value theory-based calibration of decision boundaries~\cite{6809169,7780542,7577876}. Much more work is needed in this direction, including algorithms that incorporate the risk of the unknown directly into their learning objectives, and evaluation protocols that incorporate data which is both known and unknown to a model.

\subsection*{The Role Visual Psychophysics Should Play}

One need not resort to tricky manipulations like  noise patterns that are imperceptible to humans~\cite{DBLP:journals/corr/SzegedyZSBEGF13} or carefully evolved images~\cite{7298640} to fool recognition systems based on deep learning. Simple transformations like rotation, scale, and occlusion will do the job just fine. Remarkably, a systematic study of a recognition model's performance across an exhaustive range of object appearances is typically not done during the course of machine learning research. This is a major shortcoming of evaluation within the field. Turning to the study of biological vision systems, psychologists and neuroscientists do perform such tests on humans and animals using a set of concepts and procedures from the discipline of psychophysics. Psychophysics allows scientists to probe the inner mechanisms of visual processing through the controlled manipulation of the characteristics of visual stimuli presented to a subject. The careful management of stimulus construction, ordering and presentation allows a perceptual threshold, the inflection point at which perception transitions from success to failure, to be determined precisely.  As in biological vision, we'd like to know under what conditions a machine learning model is able to operate successfully, as well as where it begins to fail. If this is to be done in an exhaustive manner, we need to leverage item response theory~\cite{embretson2000item}, which will let us map each stimulus condition to a performance point (e.g., model accuracy). When individual item responses are collected to form a curve, an exemplar-by-exemplar summary of the patterns of error for a model becomes available, allowing us to point exactly to the condition(s) that will lead to failure. 

Psychophysics is commonplace in the laboratory, but how exactly can it be applied to models? One possibility is through a computational pipeline that is able to perturb 2D natural images or 3D rendered scenes at a massive scale (e.g., millions of images per image transformation being studied) and submit them to a model, generating an item-response curve from the resulting recognition scores~\cite{DBLP:journals/corr/RichardWebsterA16}. Key to the interpretability of the results is the ability to identify a model's \textit{preferred view}. Work in vision science has established that humans possess an internalized canonical view (the visual appearance that is easiest to recognize) for individual object classes~\cite{blanz1999object}.  Similarly, recognition models have one or more preferred views of an object class, each of which leads to a maximum (or minimum) score output. A preferred view thus forms a natural starting place for model assessment. Through perturbation, the results will at best stay the same, but more likely will degrade as visual appearance moves outside the variance learned from the training dataset. With respect to the stimuli used when performing psychophysics experiments on models, there is a growing trend in robotics and computer vision to make use of simulations rendered via computer graphics. In line with this, we believe that procedurally rendered graphics hold much promise for psychophysics experiments, where the position of objects can be manipulated in 3D, and aspects of the scene, such as lighting and background, changed at will.

Instead of comparing summary statistics related to benchmark dataset performance for different models, relative performance can be assessed by comparing the respective item-response curves. Importantly, not only can any gaps between the behaviors of different models be assessed, but also potential gaps between human and model behavior. Validation by this procedure is necessary if a claim is going to made about a model matching (or exceeding) human performance. Summary statistics only reflect one data point over a mixture of scene conditions, which obscures the patterns of error we are often most interested in. Through experimentation, we have found that human performance vastly exceeds model performance even in cases where a problem has been assumed to be solved (e.g., human face detection~\cite{6701391}). While the summary statistics in those cases indicated that both humans and models were at the performance ceiling for the dataset at hand, the item-response curves from psychophysics experiments showed a clear gap between human and model performance. However, psychophysics need not entirely replace datasets. After all, we still need a collection of data from which to train the model, and some indication of performance on a collection of web-scale data is still useful for model screening. Steps should be taken to explore strategies for combining datasets and visual psychophysics to address some of the obvious shortcomings of deep learning. 
\section{The Role of Simulation for Pixel-to-Action Robotics}

Robotics, still dominated by complex processing stacks, could benefit from a similar revolution as seen in computer vision which would clear a path directly from pixels to torques and enable powerful gradient-driven end-to-end optimisation. A critical difference is that robotics constitutes an interactive domain with sequential actions where supervised learning from static datasets is not a solution. \emph{Deep reinforcement learning} is a new learning paradigm that is capable of learning end-to-end robotic control tasks, but the accomplishments have been demonstrated primarily in simulation, rather than on actual robot platforms \cite{Levine2014Guided,Schulmanetal_ICML2015,Heess2015Stochastic,Lillicrap2016Continuous,Schulmanetal_ICLR2016,mnih2016a3c,Gu2016NAF}. However, demonstrating learning capabilities on real robots remains the bar by which we must measure the practical applicability of these methods. This poses a significant challenge, given the long, data-hungry training paradigm of pixel-based deep RL methods and the relative frailty of research robots and their human handlers. 

\begin{figure}
  \centering
    \includegraphics[width=.23\textwidth]{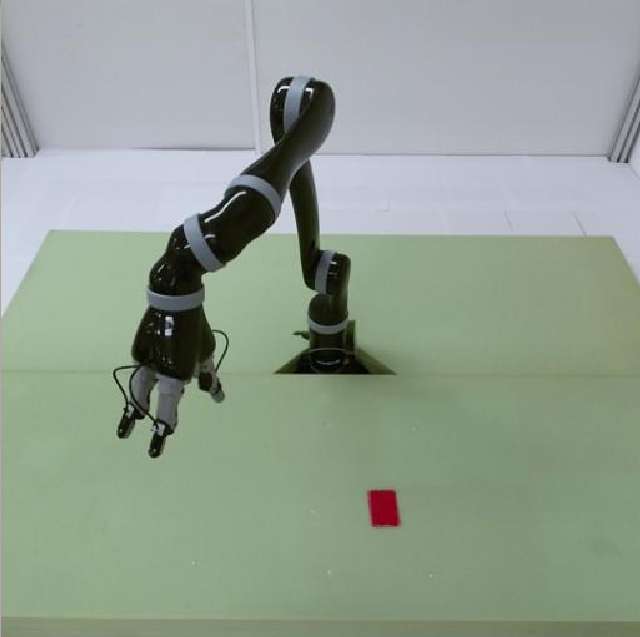}\hspace{.1in}
    \includegraphics[width=.23\textwidth]{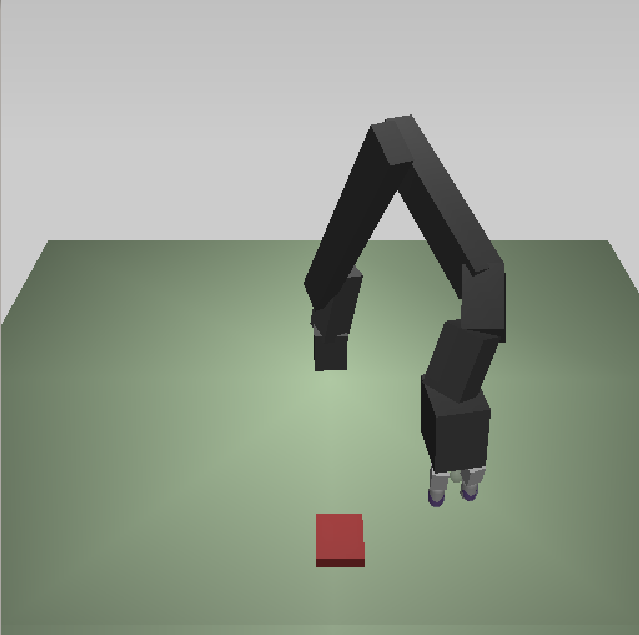}
    \caption{Sample images from the real camera input image (left) and the MuJoCo-rendered image (right), demonstrating the reality gap between simulation and reality even for a simple reaching task.
    }
    \label{fig:jacoarm}
\end{figure}

To make the challenge more concrete, consider a simple pixel-to-action learning task: reaching to a randomly placed target from a random start location, using a three-fingered Jaco robot arm (see Figure \ref{fig:jacoarm}). Trained in the MuJoCo simulator using Asynchronous Advantage Actor-Critic (A3C) \cite{mnih2016a3c}, the current state-of-the-art RL algorithm, full performance is only achieved after substantial interaction with the environment, on the order of 50 million steps - a number which is infeasible with a real robot. The simulation training, compared with the real robot, is accelerated because of fast rendering, multi-threaded learning algorithms, and the ability to continuously train without human involvement. We calculate that learning this task, which trains to convergence in 24 hours using a CPU compute cluster, would take 53 days on the real robot even with continuous training for 24 hours a day. Moreover, multiple experiments in parallel were used to explore hyperparameters in simulation; this sort of search would compound further the hypothetical real robot training time. 

Taking advantage of the simulation-learnt policies to train real robots is thus critical, but there is a \emph{reality gap} that often separates a simulated task and its real-world analogue, especially for raw pixel inputs. One solution is to use transfer learning methods to bridge the reality gap that separates simulation from real world domains.
There exist many different paradigms for domain transfer and many approaches designed specifically for deep neural models, but substantially fewer approaches for transfer from simulation to reality for robot domains. Even more rare are methods that can be used for transfer in interactive, rich sensor domains using end-to-end (pixel-to-action) learning. A growing body of work has been investigating the ability of deep networks to transfer between domains. Some research \cite{PengSAS15, SuQLG15} considers simply augmenting the target domain data with data from the source domain where an alignment exists. Building on this work, \cite{LongC0J15} starts from the observation that as one looks at higher layers in the model, the transferability of the features decreases quickly. To correct this effect, a soft constraint is added that enforces the distribution of the features to be more similar. In \cite{LongC0J15}, a `confusion' loss is proposed which forces the model to ignore variations in the data that separate the two domains \cite{TzengHDS15, TzengHZSD14}, and \cite{TzengDHFPLSD15} attempts to address the simulation to reality gap by using aligned data. The work is focused on pose estimation of the robotic arm, where training happens on a triple loss that looks at aligned simulation to real data, including the domain confusion loss. The paper does not show the efficiency of the method on learning novel complex policies. Partial success on transferring from simulation to a real robot has been reported \cite{AAMASWS10-barrett,2016JamesJohns,ZhuMKLGFF16,zhang2015towards}. They focus primarily on the problem of transfer from a more restricted simpler version of a task to the full, more difficult version. Another promising recent direction is domain randomization \cite{tobin2017domain, sadeghi2016cad2rl}.

\begin{figure}
  \centering
    \includegraphics[width=.3\textwidth]{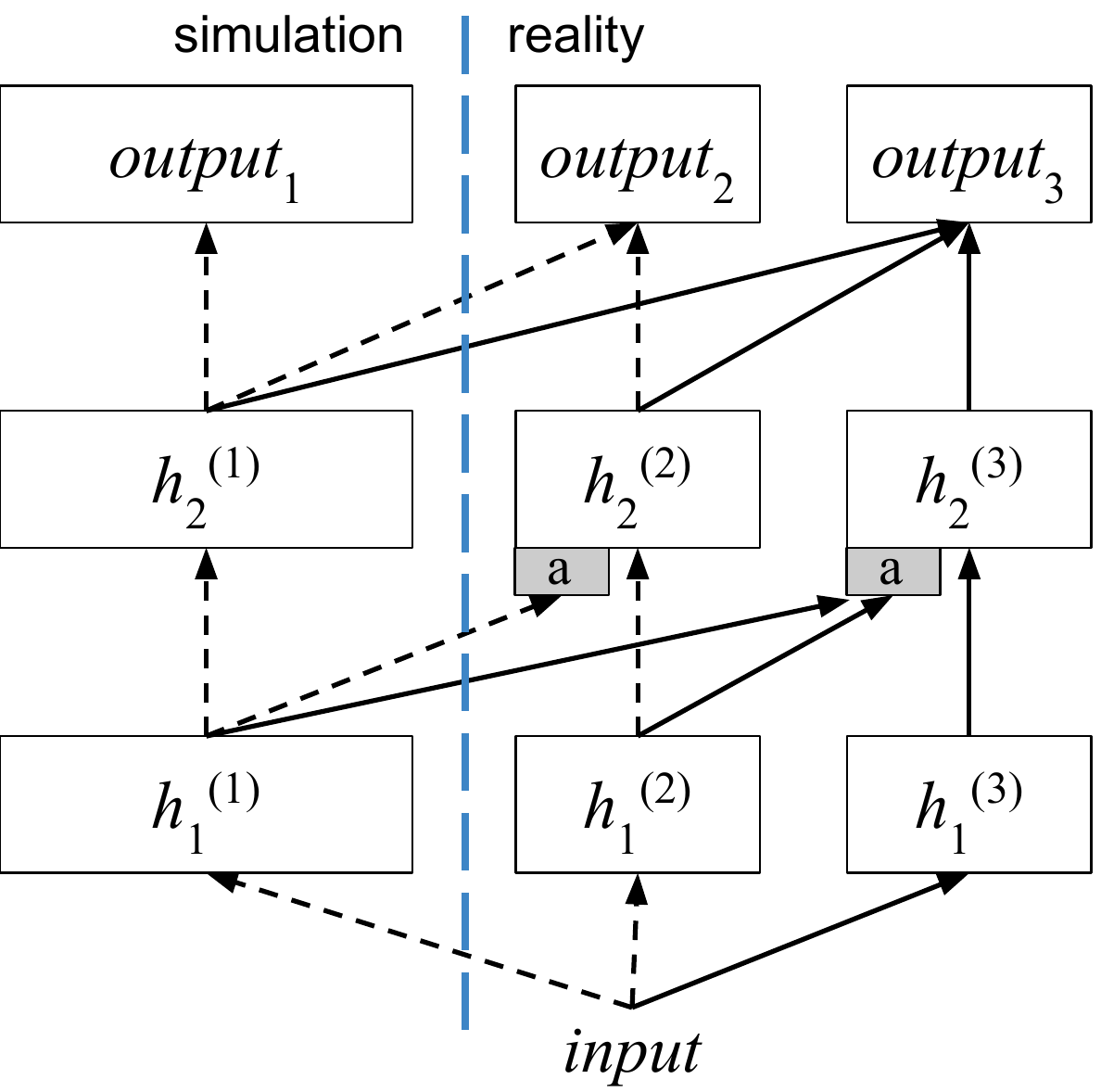}
    \caption{Detailed schematic of progressive recurrent network architecture, where the left column is trained in simulation, then the weights are frozen while the second column is trained on the real robot. A third column may then be trained on an additional task, taking advantage of the policies and features learnt and frozen in the first two columns.}
    \label{fig:networks}
\end{figure}

A recent sim-to-real approach relies on the \emph{progressive nets} architecture \cite{rusu2016progressive}, which enables transfer learning through lateral connections which connect each layer of previously learnt deep networks to new networks, thus supporting deep compositionality of features (see Figure \ref{fig:networks}). Progressive networks are well suited for sim-to-real transfer of policies in robot control domains for multiple reasons. First, features learnt for one task may be transferred to many new tasks without destruction from fine-tuning. Second, the columns may be heterogeneous, which may be important for solving different tasks, including different input modalities, or simply to improve learning speed when transferring to the real robot. Third, progressive nets add new capacity, including new input connections, when transferring to new tasks. This is advantageous for bridging the reality gap, to accommodate dissimilar inputs between simulation and real sensors.

Experiments with the Jaco robot showed that the progressive architecture is valuable for sim-to-real transfer. The progressive second column gets to 34 points, while the experiment with finetuning, which starts with the simulation-trained column and continues training on the robot, does not reach the same score as the progressive network.

\section{Deep Learning and Physics-Based Models}

\begin{table*}[t]
\centering
\caption{Models vs.~Deep Learning}
\label{tab:dl-models}
\begin{tabular}{@{}p{0.15\textwidth}p{0.4\textwidth}p{0.4\textwidth}@{}}
  \toprule
  & Model-based & Deep learning \\ \midrule
  Representation & explicit; based on or inspired by physics & implicit; network structure and parameters \\ 
  Generality & broadly applicable; physics are universal & only in trained regime; risk of overfitting  \\ 
  Robustness &  small basin of convergence; requires good models and estimates
               thereof & large basin of convergence; highly robust in trained regime \\
  Data Efficiency & very high; only needed for system identification & training requires significant data collection effort; \\ 
  Computational Efficiency & good in local regime & highly efficient once trained \\ \bottomrule
\end{tabular}
\end{table*}

The predominant approach to perception, planning, and control in
robotics is to use approximate models of the physics underlying a
robot, its sensors, and its interactions with the environment. These
model-based techniques often capture properties such as the mass,
momentum, shape, and surface friction of objects, and use these to
generate controls that change the environment in a desirable
way~\cite{todorov2012,kunze2015,Sch15Rob,Kui15Opt}.  While
physics-based models are well suited for planning and predicting the
outcome of actions, to function on a real robot they require that all
relevant model parameters are known with sufficient accuracy and can
be tracked over time. This requirement poses overly challenging
demands on system identification and perception, resulting in systems
that are brittle, especially when direct interaction with the
environment is required.

Humans, on the other hand, operate under intuitive rather than exact
physical
models~\cite{mccloskey1983,povinelli2000,Hes09Occ,Bai12Obj,Bai11How,battaglia2013simulation}. While these intuitive models have many
well-documented deficiencies and inaccuracies, they have the crucial
property that they are grounded in real world experience, are well
suited for closed-loop control, and can be learned and adapted to new
situations. As a result, humans are capable of robustly performing a
wide variety of tasks that are well beyond the reach of current robot
systems, including dexterous manipulation, handling vastly different
kinds of ingredients when cooking a meal, or climbing a tree.

Recent approaches to end-to-end training of deep networks forgo the
use of explicit physics models, learning predictive models and
controls from raw
experiences~\cite{Fin16Uns,Byravan-2016,wu2015galileo,yildirim2017physical,battaglia2016interaction,greff2017neural}.
While these early applications of large scale deep learning are just
the beginning, they have the potential to provide robots with highly
robust perception and control mechanisms, based on an intuitive notion
of physics that is fully grounded in a robot's experience.

The properties of model-based and deep learned approaches can be measured along
multiple dimensions, including the kind of representations used for reasoning,
how generally applicable their solutions are, how robust they are in real world
settings, how efficiently they make use of data, and how computationally
efficient they are during operation.  Model-based approaches often rely on
explicit models of objects and their shape, surface, and mass properties, and
use these to predict and control motion through time.  In deep learning, models
are typically implicitly encoded via networks and their parameters. As a
consequence, model-based approaches have wide applicability, since the physics
underlying them are universal.  However, at the same time, the parameters of
these models are difficult to estimate from perception, resulting in rather
brittle performance operating only in local basins of convergence. Deep learning
on the other hand enables highly robust performance when trained on sufficiently
large data sets that are representative of the operating regime of the
system. However, the implicit models learned by current DL techniques do not
have the general applicability of physics-based reasoning. Model-based
approaches are significantly more data efficient, related to their smaller
number of parameters.  The optimizations required for model-based approaches can
be performed efficiently, but the basin of convergence can be rather small.  In
contrast, deep learned solutions are often very fast and can have very large
basins of convergence. However, they do not perform well if applied in a regime
outside the training data. Table~\ref{tab:dl-models} summarizes the main
properties.

Different variants of deep learning have been shown to successfully
learn predictive physics models and robot control policies in a purely
data driven
way~\cite{Agr16Lea,Byr18Str,watter2015embed,jonschkowski2017pves}.
While such a learning-based paradigm could potentially inherit the
robustness of intuitive physics reasoning, current approaches are
nowhere near human prediction and control capabilities.  Key
challenges toward achieving highly robust, physics-based reasoning and
control for robots are: (1) Learn general, predictive models for how
the environment evolves and how it reacts to a robot’s actions. While the
first attempts in this direction show promising results, these only
capture very specific scenarios and it is not clear how they can be
made to scale to general predictive models. (2) Leverage existing
physics-based models to learn intuitive models from less data. Several
systems approach this problem in promising ways, such as using
physics-based models to generate training data for deep learning or
developing deep network structures that incorporate insights from
physics-based reasoning. (3) Learn models and controllers at multiple
levels of abstractions that can be reused in many contexts. Rather
than training new network structures for each task, such an approach
would enable robots to fully leverage previously learned knowledge and
apply it in new contexts.
\section{Towards an Automation of Informatics}

Deep learning will change the foundations of computer science.  Already, the successes of deep learning in various domains are calling into question the dominant problem-solving paradigm: algorithm design.\footnote{The term \textit{algorithm} refers to the Oxford Dictionary definition: "a process or set of rules to be followed in calculations or other problem-solving operations." Here, it includes physics formulae, computational models, probabilistic representations and inference, etc.} This can easily be seen in the area of image classification, where deep learning has outperformed all prior attempts of explicitly programming image processing algorithms.  And in contrast to most other applications of machine learning that require the careful design of problem-specific features, deep learning approaches require little to no knowledge of the problem domain.  Sure, the search for a suitable network architectures and training procedures remains but the amount of domain-specific knowledge required to apply deep learning methods to novel problem domains is substantially lower than for programming a solution explicitly.  As a result, the amount of problem-specific expertise required to solve complex problems has reached an all-time low. Whether this is good or bad remains to be seen (it is probably neither and both). But it might seem that deep learning is currently the winner in the competition between ``traditional" \textit{programming} and the clever use of large amounts of \textit{data}.

\subsection*{Programming versus data}

Solutions to computational problems lie on a spectrum along which the relative and complementary contributions of programming and data vary. On one end of the spectrum lies traditional computer science: human experts program problem-specific algorithms that require no additional data to solve a particular problem instance, e.g. quicksort. On the other extreme lies deep learning.  A generic approach to learning leverages large amounts of data to find a computational solution automatically.  In between these two extremes lie algorithms that are less generic than deep learning and less specific than quicksort, including maybe decision trees for example.

It is helpful to look at the two ends of the spectrum in more detail.  The act of \textit{programming} on one end of the spectrum is replaced by \textit{training} on the other end.  The concept of \textit{program} is turned into \textit{learning weights} of the network.  And the programming language, i.e.~the language in which a solution is expressed, is replaced by network architecture, loss function, training procedure, and data.  Please note that the training procedure itself is again seen as a concrete algorithm, on the opposing end of the spectrum.  This already alludes to the fact that solutions to challenging problems probably must combine sub-solutions from the entire spectrum spanned by programming and deep learning.

\subsection*{Does understand imply one end of the spectrum?}

For a programmer to solve a problem through programming, we might say that she has to \textit{understand} the problem.  Computer programs therefore reflect human understanding.  We might also say that the further a particular solution is positioned towards the deep-learning-end of the spectrum, the less understanding about the problem it requires.  As science strives for understanding, we should ultimately attempt to articulate the structure of our solutions explicitly, relying on as little data as possible for solving a particular problem instance. There are many reasons for pursuing this goal: robustness, transfer, generality, verifiability, re-use, and ultimately insight, which might lead to further progress.

Consider, for example, the problem of tracking the trajectory of a quad-copter. We can certainly come up with a deep learning solution to this problem.  But would we not expect the outcome of learning, given an arbitrary amount of  data and computational resources, to be some kind of Bayes filter?  Either we believe that the Bayes filter captures the computational structure inherent to this problem (recursive state estimation), and then a learned solution eventually has to discover and represent this solution.  But at that point we might simply use the algorithm instead of the deep neural network.  If, on the other hand, the deep neural network represents something else than a Bayes filter---something outperforming the Bayes filter---then we discovered that Bayes filters do not adequately capture the structure of the problem at hand. And we will naturally be curious as to what the neural network discovered.

From this, we should draw three conclusions: First, our quest for understanding implies that we must try to move towards the programming-end of the spectrum, whenever we can.  Second, we need to be able to leverage generic tools, such as deep learning, to discover problem structure; this will help us derive novel knowledge and to devise algorithms based on that knowledge.  Third, we should understand how problems can be divided into parts: those parts for which we know the structure (and therefore can write algorithms for) and those for which we would like to discover the structure. This will facilitate the component-wise movement towards explicit understanding.

\subsection*{Generic tools might help us identify new structure}

When we do not know how to program a solution for a problem and instead apply a generic learning method, such as deep learning, and this generic method delivers a solution, then we have implicitly learned something about the problem.  It might be difficult to extract this knowledge from a deep neural network but that should simply motivate us to develop methods for extracting this knowledge. Towards this goal, our community should \textit{a)} report in detail on the limitations of deep networks and \textit{b)} study in similar detail the dependencies of deep learning solutions on various parameters. This will lead the way to an ability of ``reading'' networks so as to extract algorithmifiable information. 

There have been some recent results about ``distilling'' knowledge from neural networks, indicating that the extraction of problem structure from neural networks might be possible~\cite{Hinton-2015}.  Such distilled knowledge is still far away from being algorithmifiable, but this line of work seem promising in this regard.  The idea of distillation can also be combined with side information~\cite{Lopez-Paz-2016}, further facilitating the identification of relevant problem structure.

On the other hand, it was shown that our insights about generalization---an important objective for machine learning algorithms---might not transfer easily to neural networks~\cite{Zhang-2016}.  If it turns out that the deep neural networks we learn today simply memorize training data and then interpolate between them~\cite{Zhang-2016}, then we must develop novel regularization methods to enforce the extraction of problem structure instead of memorization, possibly through the use of side information~\cite{Jonschkowski-2015side}.  Or, if neural networks are only good for memorization, they are not as powerful as we thought. There might be evidence, however, that neural networks do indeed find good representations, i.e.~problem structure.

\subsection*{Complex problems should be solved by decomposition and re-composition}

In many cases, interesting and complex problems will exhibit complex structure because they are composed of sub-problems. For each of these sub-problems, computational solutions are most appropriate that lie on different points along the programming/data-spectrum; this is because we may have more or less understanding of the sub-problem's inherent structure. It would therefore make sense to compose solutions to the original problem from sub-solutions that lie on different points on the programming/data-spectrum~\cite{Jonschkowski-2016}. 

For many sub-problems, we already have excellent algorithmic solutions, e.g.~implementations of quicksort.  Sorting is a problem on one end of the spectrum: we understand it and have codified that understanding in an algorithm.  But there are many other problems, such as image classification, where human programs are outperformed by deep neural networks. Those problem should be solved by neural networks and then integrated with solutions from other parts of the spectrum.

This re-composition of component solutions from different places on the spectrum can be achieved with differentiable versions of existing algorithms (one end of the spectrum) that are compatible solutions obtained with back-propagation (other end of the spectrum)~\cite{Shi-2009, Wilson-2009, Byravan-2016, Haarnoja-2016, Tamar-2016, Jonschkowski-2016}.  For example, Jonschkowski~et~al.~\cite{Jonschkowski-2016} solve the aforementioned localization problem for quad-copters by combining a histogram filter with back-propagation-learned motion and sensing models.

\subsection*{Decomposability of problems}

In the previous section, I argued that complex problems often are decomposable into sub-problems that can be solved independently.  A problem is called \textit{decomposable} or \textit{near-decomposable}~\cite{Simon-96} if there is little complexity in the interactions among sub-problems and most of the complexity is handled within those sub-problems.  But are all problems decomposable in this manner?  For example, Schierwagen argued that the brain is not decomposable~\cite{schierwagen_reverse_2012} because the interactions between its components still contain much of the complexity of the original problem.  Furthermore, many interpret results on end-to-end learning of deep visuomotor policies to indicate that modular sub-solutions automatically lead to poor solutions~\cite{Levine-2016}.  Of course, a sub-optimal factorization of a problem into sub-problems will lead to sub-optimal solutions.  However, the results presented by Levine et al.~\cite{Levine-2016} do not lend strong support to this statement. The authors show that end-to-end learning, i.e.\ giving up strict boundaries between sub-problems improves their solution. However, it is unclear if this is an artifacts of overfitting, an indication of a poor initial factorization, or an indication of the fact that even correct factorizations may exclude parts of the solution space containing the optimal solution.

Irrespective of the degree of decomposability of a problem (and the suitable degree of modularity of the solution), we suspect that there are optimal factorizations of problems for a defined task, agent, and environment.  Such a factorization may not always lead to simple interfaces between sup-problems but always facilitates finding an optimal solution.

\subsection*{Automating programming}

Once we are able to \textit{1)}~decompose problems into sub-problems, \textit{2)}~solve those sub-problems with solutions from different points along the programming/data-spectrum, \textit{3)}~recompose the solutions to sub-problems, and \textit{4)}~extract algorithmic information from data-driven solutions, we might as well automate programming (computer science?) altogether.  Programming should be easy to automate, as it takes place entirely within the well-defined world of the computer.  If we can successfully apply generic methods to complex problems, extract and algorithmify structural knowledge from the resulting solutions, use the resulting algorithms to solve sub-problems of the original problem, thereby making that original problem more easily solvable, and so forth---then we can also imagine an automated way of deriving computer algorithms from problem-specific data.  A key challenge will be the automatic decomposition or \textit{factorization} of the problem into suitably solvable sub-problems.  

This view raises some fundamental questions about the differences between \textit{program} in programming and \textit{weights} in deep learning.  Really, this view implies that there is no qualitative difference between them, only a difference of expressiveness and the amount of prior assumptions reflected in them. Programs and weights, in this view, are different instances of the same thing, namely of parameters that specify a solution, given a framework for expressing such solutions.  Now it seems plausible that we can incrementally extract structure from learned parameters (weights), leading to a less generic representation with fewer parameters, until the parameters are so specific that we might call them a program.

But the opposite is also possible.  It is possible that problems exists that do not exhibit algorithmifiable structure.  And it is possible that these problems can (only) be solved in a data-driven manner.  To speculate about this, comparisons with biological cognitive capabilities might be helpful: Can these capabilities (in principle) be encoded in a program?  Do these capabilities depend on massive amounts of data? These are difficult questions that AI researchers have asked themselves for many years.

\subsection*{Priors to reduce the amount of data}

A natural concern for this kind of reasoning is the necessity to acquire large amounts of data. This can be very costly, especially when this data has to be acquired from interaction with the real world, as it is the case in robotics.  It will then become necessary to reduce the required amount of data by incorporating appropriate priors into learning~\cite{Jonschkowski-2015}.  These priors reduce all possible interpretations of data to only those consistent with the prior.  If sufficiently strong priors are available, it will become possible to extract (and possibly algorithmify) the problem structure from reasonable amounts of data.

It might also be difficult to separate acquired data into those groups associated with a single task. Recent methods have shown that this separation can be performed automatically~\cite{Hoefer-2016}. Now data can be acquired in less restrictive settings and the learning agent can differentiate the task associated with a datum by itself.

\subsection*{Where will this lead?}

Maybe in the end, the most lasting impact of deep learning will not be deep learning itself but rather the effect it had.  The successes of deep learning, achieved by leveraging data and computation, have made computer scientists realize that there is a spectrum---rather than a dichotomy---between programming and data. This realization may pave the way for a computer science that fully leverages the entire breadth of this spectrum to automatically derive algorithms from reasonable amounts of data and suitable priors.

\section{Conclusions}
The rather skeptical attitude towards deep learning at the \emph{Robotics: Science and Systems} (RSS) conference in Rome 2015 motivated us to organize a workshop at RSS 2016 with the title \emph{``Are the Skeptics Right? Limits and Potentials of Deep Learning in Robotics''}~\cite{rss16-ws-skeptics}. As it turned out, by then there were hardly any skeptics left. The robotics community had accepted deep learning as a very powerful tool and begun to utilize and advance it. A follow-up workshop on \emph{``New Frontiers for Deep Learning in Robotics''}~\cite{rss17-ws-deepLearning} at RSS 2017 concentrated more on some of the robotics-specific research challenges we discussed in this paper.
2017 saw a surge of deep learning in robotics:  workshops at CVPR~\cite{cvpr17-ws-roboticVision} and NIPS~\cite{nips17-ws-robots} built bridges between the robotics, computer vision, and machine learning communities. Over 10\% of the papers submitted to ICRA 2018 used \emph{Deep learning in robotics and automation} as a keyword, making it the most frequent keyword. Furthermore, a whole new Conference on Robot Learning (CoRL)~\cite{corl} was initiated. 

While much ongoing work in deep learning for robotics concentrates on either perception or acting, we hope to see more integrated approaches in the future: robots that learn to utilize their embodiment to reduce the uncertainty in perception, decision making, and execution. Robots that learn complex  multi-stage tasks, while incorporating prior model knowledge or heuristics, and exploiting a semantic understanding of their environment. Robots that learn to discover and exploit the rich semantic regularities and geometric structure of the world, to operate more robustly in realistic environments with open-set characteristics.

Deep learning techniques have revolutionized many aspects of computer vision over the past five years and have  been rapidly adopted into robotics as well. However, robotic perception, robotic learning, and robotic control are demanding tasks that continue to pose severe challenges on the techniques typically applied. 
Our paper discussed some of these current research questions and challenges for deep learning in robotics.
We pointed the reader into different directions worthwhile for further research and hope our paper contributes to the ongoing advancement of deep learning for robotics.

\section*{Acknowledgements}
This work was supported by the Australian Research Council Centre of Excellence for Robotic Vision, project number CE140100016. Oliver Brock was supported by DFG grant 329426068. Walter Scheirer acknowledges the funding provided by IARPA contract D16PC00002. Michael Milford was partially supported by an Australian Research Council Future Fellowship FT140101229.

\bibliographystyle{plain}
\bibliography{raia,oliver,dieter,walter,niko,workshops}

\begin{thebibliography}{100}

\bibitem{corl}
{Conference on Robot Learning (CoRL)}.
\newblock \url{http://www.robot-learning.org}, 2017.

\bibitem{Agr16Lea}
P.~Agrawal, AV. Nair, P.~Abbeel, J.~Malik, and S.~Levine.
\newblock Learning to poke by poking: Experiential learning of intuitive
  physics.
\newblock In {\em {Advances in Neural Information Processing Systems (NIPS)}},
  2016.

\bibitem{cvpr17-ws-roboticVision}
Anelia Angelova, Gustavo Carneiro, Kevin Murphy, Niko S\"underhauf, J\"urgen
  Leitner, Ian Lenz, Trung~T. Pham, Vijay Kumar, Ingmar Posner, Michael
  Milford, Wolfram Burgard, Ian Reid, and Peter Corke.
\newblock {Computer Vision and Pattern Recognition (CVPR) Workshop on Deep
  Learning in Robotic Vision}.
\newblock
  \url{http://juxi.net/workshop/deep-learning-robotic-vision-cvpr-2017/}, 2017.

\bibitem{anthony}
Samuel~E. Anthony.
\newblock The trollable self-driving car.
\newblock Slate, March 2016.
\newblock Accessed 2016-12-21 via \url{http://goo.gl/78fglb}.

\bibitem{atanasov2014nonmyopic}
Nikolay Atanasov, Bharath Sankaran, Jerome Le~Ny, George~J Pappas, and Kostas
  Daniilidis.
\newblock Nonmyopic view planning for active object classification and pose
  estimation.
\newblock {\em IEEE Transactions on Robotics}, 30(5):1078--1090, 2014.

\bibitem{Bai11How}
R.~Baillargeon, J.~Li, Y.~Gertner, and D.~Wu.
\newblock How do infants reason about physical events?
\newblock In {\em The Wiley-Blackwell handbook of childhood cognitive
  development, second edition}. Oxford: Blackwell, 20111.

\bibitem{Bai12Obj}
R.~Baillargeon, M.~Stavans, D.~Wu, R.~Gertner, P.~Setoh, A.~K. Kittredge, and
  A.~Bernard.
\newblock Object individuation and physical reasoning in infancy: An
  integrative account.
\newblock {\em Language Learning and Development}, 8, 2012.

\bibitem{AAMASWS10-barrett}
Samuel Barrett, Matthew~E. Taylor, and Peter Stone.
\newblock Transfer learning for reinforcement learning on a physical robot.
\newblock In {\em Ninth International Conference on Autonomous Agents and
  Multiagent Systems - Adaptive Learning Agents Workshop (AAMAS - ALA)}, 2010.

\bibitem{battaglia2016interaction}
Peter Battaglia, Razvan Pascanu, Matthew Lai, Danilo~Jimenez Rezende, et~al.
\newblock Interaction networks for learning about objects, relations and
  physics.
\newblock In {\em Advances in neural information processing systems}, pages
  4502--4510, 2016.

\bibitem{battaglia2013simulation}
Peter~W Battaglia, Jessica~B Hamrick, and Joshua~B Tenenbaum.
\newblock Simulation as an engine of physical scene understanding.
\newblock {\em Proceedings of the National Academy of Sciences},
  110(45):18327--18332, 2013.

\bibitem{bendale2015towards}
Abhijit Bendale and Terrance~E. Boult.
\newblock Towards open world recognition.
\newblock In {\em Proceedings of the IEEE Conference on Computer Vision and
  Pattern Recognition}, pages 1893--1902, June 2015.

\bibitem{7780542}
Abhijit Bendale and Terrance~E. Boult.
\newblock Towards open set deep networks.
\newblock In {\em Proceedings of the IEEE Conference on Computer Vision and
  Pattern Recognition}, pages 1563--1572, June 2016.

\bibitem{bertinetto2016learning}
Luca Bertinetto, Jo{\~a}o~F Henriques, Jack Valmadre, Philip Torr, and Andrea
  Vedaldi.
\newblock Learning feed-forward one-shot learners.
\newblock In {\em Advances in Neural Information Processing Systems}, pages
  523--531, 2016.

\bibitem{bircher2016receding}
Andreas Bircher, Mina Kamel, Kostas Alexis, Helen Oleynikova, and Roland
  Siegwart.
\newblock Receding horizon" next-best-view" planner for 3d exploration.
\newblock In {\em IEEE International Conference on Robotics and Automation
  (ICRA)}, pages 1462--1468. IEEE, 2016.

\bibitem{blanz1999object}
Volker Blanz, Michael~J. Tarr, and Heinrich~H. B{\"u}lthoff.
\newblock What object attributes determine canonical views?
\newblock {\em Perception}, 28(5):575--599, 1999.

\bibitem{Byravan-2016}
A.~Byravan and D.~Fox.
\newblock {SE3}-nets: Learning rigid body motion using deep neural networks.
\newblock In {\em Proc.~of the IEEE International Conference on Robotics \&
  Automation (ICRA)}, 2017.

\bibitem{Byr18Str}
A.~Byravan, F.~Leeb, F.~Meier, and D.~Fox.
\newblock {SE3-Pose-Nets}: Structured deep dynamics models for visuomotor
  planning and control.
\newblock In {\em {Proc.~of the IEEE International Conference on Robotics \&
  Automation (ICRA)}}, 2018.

\bibitem{Fin16Uns}
I.~Goodfellow C.~Finn and S.~Levine.
\newblock Unsupervised learning for physical interaction through video
  prediction.
\newblock In {\em {Advances in Neural Information Processing Systems (NIPS)}},
  2016.

\bibitem{cadena2016past}
Cesar Cadena, Luca Carlone, Henry Carrillo, Yasir Latif, Davide Scaramuzza,
  Jos{\'e} Neira, Ian Reid, and John~J Leonard.
\newblock Past, present, and future of simultaneous localization and mapping:
  Toward the robust-perception age.
\newblock {\em IEEE Transactions on Robotics}, 32(6):1309--1332, 2016.

\bibitem{choy20163d}
Christopher~B Choy, Danfei Xu, JunYoung Gwak, Kevin Chen, and Silvio Savarese.
\newblock 3d-r2n2: A unified approach for single and multi-view 3d object
  reconstruction.
\newblock In {\em European Conference on Computer Vision (ECCV)}, pages
  628--644. Springer, 2016.

\bibitem{cohn1996active}
David~A Cohn, Zoubin Ghahramani, and Michael~I Jordan.
\newblock Active learning with statistical models.
\newblock {\em Journal of artificial intelligence research}, 4(1):129--145,
  1996.

\bibitem{cox2014neural}
David~D. Cox and Thomas Dean.
\newblock Neural networks and neuroscience-inspired computer vision.
\newblock {\em Current Biology}, 24(18):R921--R929, 2014.

\bibitem{csurka2017domain}
Gabriela Csurka.
\newblock Domain adaptation for visual applications: A comprehensive survey.
\newblock {\em arXiv preprint arXiv:1702.05374}, 2017.

\bibitem{dayoub2017episode}
Feras Dayoub, Niko S{\"u}nderhauf, and Peter Corke.
\newblock Episode-based active learning with bayesian neural networks.
\newblock In {\em CVPR Workshop on Deep Learning for Robotic Vision. arXiv
  preprint arXiv:1703.07473}, 2017.

\bibitem{doumanoglou2016recovering}
Andreas Doumanoglou, Rigas Kouskouridas, Sotiris Malassiotis, and Tae-Kyun Kim.
\newblock Recovering 6d object pose and predicting next-best-view in the crowd.
\newblock In {\em Proceedings of the IEEE Conference on Computer Vision and
  Pattern Recognition}, pages 3583--3592, 2016.

\bibitem{embretson2000item}
Susan~E. Embretson and Steven~P. Reise.
\newblock {\em Item response theory for psychologists}.
\newblock Lawrence Erlbaum Associates, Inc., 2000.

\bibitem{finn2017model}
Chelsea Finn, Pieter Abbeel, and Sergey Levine.
\newblock Model-agnostic meta-learning for fast adaptation of deep networks.
\newblock {\em arXiv preprint arXiv:1703.03400}, 2017.

\bibitem{gal2016dropout}
Yarin Gal and Zoubin Ghahramani.
\newblock Dropout as a bayesian approximation: Representing model uncertainty
  in deep learning.
\newblock In {\em International Conference on Machine Learning (ICML)}, pages
  1050--1059, 2016.

\bibitem{gal2017deep}
Yarin Gal, Riashat Islam, and Zoubin Ghahramani.
\newblock Deep bayesian active learning with image data.
\newblock {\em arXiv preprint arXiv:1703.02910}, 2017.

\bibitem{Ganin2015domain}
Yaroslav Ganin, Evgeniya Ustinova, Hana Ajakan, Pascal Germain, Hugo
  Larochelle, Fran{\c{c}}ois Laviolette, Mario Marchand, and Victor~S.
  Lempitsky.
\newblock Domain-adversarial training of neural networks.
\newblock {\em CoRR}, abs/1505.07818, 2015.

\bibitem{garg2016unsupervised}
Ravi Garg, Gustavo Carneiro, and Ian Reid.
\newblock Unsupervised cnn for single view depth estimation: Geometry to the
  rescue.
\newblock In {\em European Conference on Computer Vision}, pages 740--756.
  Springer, 2016.

\bibitem{godard2017unsupervised}
Cl{\'e}ment Godard, Oisin Mac~Aodha, and Gabriel~J Brostow.
\newblock Unsupervised monocular depth estimation with left-right consistency.
\newblock In {\em Computer Vision and Pattern Recognition (CVPR)}, 2017.

\bibitem{goldstein2016sensation}
E~Bruce Goldstein and James Brockmole.
\newblock {\em Sensation and perception}.
\newblock Cengage Learning, 2016.

\bibitem{goodfellow2013empirical}
Ian~J Goodfellow, Mehdi Mirza, Da~Xiao, Aaron Courville, and Yoshua Bengio.
\newblock An empirical investigation of catastrophic forgetting in
  gradient-based neural networks.
\newblock {\em arXiv preprint arXiv:1312.6211}, 2013.

\bibitem{goodfellow2014explaining}
Ian~J Goodfellow, Jonathon Shlens, and Christian Szegedy.
\newblock Explaining and harnessing adversarial examples.
\newblock {\em arXiv preprint arXiv:1412.6572}, 2014.

\bibitem{greff2017neural}
Klaus Greff, Sjoerd van Steenkiste, and J{\"u}rgen Schmidhuber.
\newblock Neural expectation maximization.
\newblock In {\em Advances in Neural Information Processing Systems}, pages
  6694--6704, 2017.

\bibitem{Gu2016NAF}
Shixiang Gu, Timothy~P. Lillicrap, Ilya Sutskever, and Sergey Levine.
\newblock Continuous deep q-learning with model-based acceleration.
\newblock In {\em ICML 2016}, 2016.

\bibitem{guo2017calibration}
Chuan Guo, Geoff Pleiss, Yu~Sun, and Kilian~Q Weinberger.
\newblock On calibration of modern neural networks.
\newblock {\em arXiv preprint arXiv:1706.04599}, 2017.

\bibitem{Haarnoja-2016}
Tuomas Haarnoja, Anurag Ajay, Sergey Levine, and Pieter Abbeel.
\newblock Backprop {KF:} learning discriminative deterministic state
  estimators.
\newblock {\em CoRR}, abs/1605.07148, 2016.

\bibitem{hane2017hierarchical}
Christian H{\"a}ne, Shubham Tulsiani, and Jitendra Malik.
\newblock Hierarchical surface prediction for 3d object reconstruction.
\newblock {\em arXiv preprint arXiv:1704.00710}, 2017.

\bibitem{hariharan2016low}
Bharath Hariharan and Ross Girshick.
\newblock Low-shot visual recognition by shrinking and hallucinating features.
\newblock {\em arXiv preprint arXiv:1606.02819}, 2016.

\bibitem{he2017mask}
Kaiming He, Georgia Gkioxari, Piotr Doll{\'a}r, and Ross Girshick.
\newblock Mask r-cnn.
\newblock In {\em IEEE International Conference on Computer Vision (ICCV)},
  2017.

\bibitem{Heess2015Stochastic}
Nicolas Heess, Gregory Wayne, David Silver, Timothy~P. Lillicrap, Tom Erez, and
  Yuval Tassa.
\newblock Learning continuous control policies by stochastic value gradients.
\newblock In {\em Advances in Neural Information Processing Systems 28: Annual
  Conference on Neural Information Processing Systems 2015, December 7-12,
  2015, Montreal, Quebec, Canada}, pages 2944--2952, 2015.

\bibitem{hendrycks2017baseline}
Dan Hendrycks and Kevin Gimpel.
\newblock A baseline for detecting misclassified and out-of-distribution
  examples in neural networks.
\newblock In {\em International Conference on Machine Learning (ICML)}, 2017.

\bibitem{Hes09Occ}
S.~Hespos, G.~Gredeba, C.~von Hofsten, and E.~Spelke.
\newblock Occlusion is hard: Comparing predictive reaching for visible and
  hidden objects in infants and adults.
\newblock {\em Cognitive Science}, 2009.

\bibitem{Hinton-2015}
Geoffrey Hinton, Oriol Vinyals, and Jeff Dean.
\newblock Distilling the knowledge in a neural network.
\newblock {\em CoRR}, abs/1503.02531, 2015.

\bibitem{Hoefer-2016}
Sebastian H\"ofer, Antonin Raffin, Rico Jonschkowski, Oliver Brock, and Freek
  Stulp.
\newblock Unsupervised learning of state representations for multiple tasks.
\newblock In {\em Deep Learning Workshop at the Conference on Neural
  Information Processing Systems (NIPS)}, 2016.

\bibitem{2016JamesJohns}
S.~{James} and E.~{Johns}.
\newblock {3D Simulation for Robot Arm Control with Deep Q-Learning}.
\newblock {\em ArXiv e-prints}, 2016.

\bibitem{Jonschkowski-2015}
Rico Jonschkowski and Oliver Brock.
\newblock Learning state representations with robotic priors.
\newblock {\em Autonomous Robots}, 39(3):407--428, 2015.

\bibitem{Jonschkowski-2016}
Rico Jonschkowski and Oliver Brock.
\newblock End-to-end learnable histogram filters.
\newblock In {\em Workshop on Deep Learning for Action and Interaction at the
  Conference on Neural Information Processing Systems (NIPS)}, 2016.

\bibitem{jonschkowski2017pves}
Rico Jonschkowski, Roland Hafner, Jonathan Scholz, and Martin Riedmiller.
\newblock Pves: Position-velocity encoders for unsupervised learning of
  structured state representations.
\newblock {\em arXiv preprint arXiv:1705.09805}, 2017.

\bibitem{Jonschkowski-2015side}
Rico Jonschkowski, Sebastian H{\"{o}}fer, and Oliver Brock.
\newblock Contextual learning.
\newblock {\em CoRR}, abs/1511.06429, 2015.

\bibitem{Kaess12}
M.~Kaess, H.~Johannsson, R.~Roberts, V.~Ila, J.~Leonard, and F.~Dellaert.
\newblock {iSAM2: Incremental Smoothing and Mapping Using the Bayes Tree}.
\newblock {\em Intl. Journal of Robotics Research}, 31(2):216--235, February
  2012.

\bibitem{kendall2017uncertainties}
Alex Kendall and Yarin Gal.
\newblock What uncertainties do we need in bayesian deep learning for computer
  vision?
\newblock {\em arXiv preprint arXiv:1703.04977}, 2017.

\bibitem{kersten2004object}
Daniel Kersten, Pascal Mamassian, and Alan Yuille.
\newblock Object perception as bayesian inference.
\newblock {\em Annu. Rev. Psychol.}, 55:271--304, 2004.

\bibitem{kingma2014semi}
Diederik~P Kingma, Shakir Mohamed, Danilo~Jimenez Rezende, and Max Welling.
\newblock Semi-supervised learning with deep generative models.
\newblock In {\em Advances in Neural Information Processing Systems}, pages
  3581--3589, 2014.

\bibitem{Kui15Opt}
S.~Kuindersma, R.~Deits, M.~Fallon, A.~Valenzuela, H.~Dai, F.~Permenter,
  T.~Koolen, P.~Marion, and R.~Tedrake.
\newblock Optimization-based locomotion planning, estimation, and control
  design for the {A}tlas humanoid robot.
\newblock {\em Autonomous Robots}, 40(3), 2016.

\bibitem{Kummerle11}
R.~K{\"u}mmerle, G.~Grisetti, H.~Strasdat, K.~Konolige, and W.~Burgard.
\newblock {g2o: A General Framework for Graph Optimization}.
\newblock In {\em Proc. of Intl. Conf. on Robotics and Automation (ICRA)},
  pages 3607 -- 3613, 2011.

\bibitem{kunze2015}
Lars Kunze and Michael Beetz.
\newblock Envisioning the qualitative effects of robot manipulation actions
  using simulation-based projections.
\newblock {\em Artificial Intelligence}, 2015.

\bibitem{lake2015human}
Brenden~M Lake, Ruslan Salakhutdinov, and Joshua~B Tenenbaum.
\newblock Human-level concept learning through probabilistic program induction.
\newblock {\em Science}, 350(6266):1332--1338, 2015.

\bibitem{lakshminarayanan2017simple}
Balaji Lakshminarayanan, Alexander Pritzel, and Charles Blundell.
\newblock Simple and scalable predictive uncertainty estimation using deep
  ensembles.
\newblock In {\em Advances in Neural Information Processing Systems}, pages
  6393--6395, 2017.

\bibitem{Levine2014Guided}
Sergey Levine and Pieter Abbeel.
\newblock Learning neural network policies with guided policy search under
  unknown dynamics.
\newblock In Z.~Ghahramani, M.~Welling, C.~Cortes, N.~D. Lawrence, and K.~Q.
  Weinberger, editors, {\em Advances in Neural Information Processing Systems
  27}, pages 1071--1079. Curran Associates, Inc., 2014.

\bibitem{Levine-2016}
Sergey Levine, Chelsea Finn, Trevor Darrell, and Pieter Abbeel.
\newblock End-to-end training of deep visuomotor policies.
\newblock {\em The Journal of Machine Learning Research}, 17(1):1334--1373,
  2015.

\bibitem{Lillicrap2016Continuous}
Timothy~P. Lillicrap, Jonathan~J. Hunt, Alexander Pritzel, Nicolas Heess, Tom
  Erez, Yuval Tassa, David Silver, and Daan Wierstra.
\newblock Continuous control with deep reinforcement learning.
\newblock {\em CoRR}, abs/1509.02971, 2015.

\bibitem{lin2013holistic}
Dahua Lin, Sanja Fidler, and Raquel Urtasun.
\newblock Holistic scene understanding for 3d object detection with rgbd
  cameras.
\newblock In {\em Proceedings of the IEEE International Conference on Computer
  Vision}, pages 1417--1424, 2013.

\bibitem{liu2016learning}
Fayao Liu, Chunhua Shen, Guosheng Lin, and Ian Reid.
\newblock Learning depth from single monocular images using deep convolutional
  neural fields.
\newblock {\em IEEE transactions on pattern analysis and machine intelligence},
  38(10):2024--2039, 2016.

\bibitem{liu2016ssd}
Wei Liu, Dragomir Anguelov, Dumitru Erhan, Christian Szegedy, Scott Reed,
  Cheng-Yang Fu, and Alexander~C Berg.
\newblock Ssd: Single shot multibox detector.
\newblock In {\em European conference on computer vision}, pages 21--37.
  Springer, 2016.

\bibitem{NYT5}
Steve Lohr.
\newblock A lesson of {Tesla} crashes? computer vision can't do it all yet.
\newblock The New York Times, September 2016.
\newblock Accessed 2016-12-21 via \url{http://goo.gl/5RcHVr}.

\bibitem{lomonaco2017core50}
Vincenzo Lomonaco and Davide Maltoni.
\newblock Core50: a new dataset and benchmark for continuous object
  recognition.

\bibitem{LongC0J15}
Mingsheng Long, Yue Cao, Jianmin Wang, and Michael~I. Jordan.
\newblock Learning transferable features with deep adaptation networks.
\newblock In {\em Proceedings of the 32nd International Conference on Machine
  Learning, {ICML} 2015, Lille, France, 6-11 July 2015}, pages 97--105, 2015.

\bibitem{Lopez-Paz-2016}
David Lopez-Paz, L\'eon Bottou, Bernhard Sch\"olkopf, and Vladimir Vapnik.
\newblock Unifying distillation and privileged information.
\newblock {\em CoRR}, abs/1511.03643, 2016.

\bibitem{mackay1992practical}
David~JC MacKay.
\newblock A practical bayesian framework for backpropagation networks.
\newblock {\em Neural computation}, 4(3):448--472, 1992.

\bibitem{malmir2017deep}
Mohsen Malmir, Karan Sikka, Deborah Forster, Ian Fasel, Javier~R Movellan, and
  Garrison~W Cottrell.
\newblock Deep active object recognition by joint label and action prediction.
\newblock {\em Computer Vision and Image Understanding}, 156:128--137, 2017.

\bibitem{mccloskey1983}
Michael McCloskey.
\newblock Intuitive physics.
\newblock {\em Scientific american}, 248(4):122--130, 1983.

\bibitem{mensink2012metric}
Thomas Mensink, Jakob Verbeek, Florent Perronnin, and Gabriela Csurka.
\newblock Metric learning for large scale image classification: Generalizing to
  new classes at near-zero cost.
\newblock {\em Computer Vision--ECCV 2012}, pages 488--501, 2012.

\bibitem{miller2017dropout}
Dimity Miller, Lachlan Nicholson, Feras Dayoub, and Niko S{\"u}nderhauf.
\newblock Dropout sampling for robust object detection in open-set conditions.
\newblock In {\em International Conference on Robotics and Automation (ICRA)},
  2017.

\bibitem{mnih2016a3c}
Volodymyr Mnih, Adrià~Puigdomènech Badia, Mehdi Mirza, Alex Graves,
  Timothy~P. Lillicrap, Tim Harley, David Silver, and Koray Kavukcuoglu.
\newblock Asynchronous methods for deep reinforcement learning.
\newblock In {\em Int'l Conf. on Machine Learning (ICML)}, 2016.

\bibitem{neal1995bayesian}
Radford~M Neal.
\newblock {\em Bayesian learning for neural networks}.
\newblock PhD thesis, University of Toronto, 1995.

\bibitem{nguyen2015deep}
Anh Nguyen, Jason Yosinski, and Jeff Clune.
\newblock Deep neural networks are easily fooled: High confidence predictions
  for unrecognizable images.
\newblock In {\em Proceedings of the IEEE Conference on Computer Vision and
  Pattern Recognition}, pages 427--436, 2015.

\bibitem{7298640}
Anh Nguyen, Jason Yosinski, and Jeff Clune.
\newblock Deep neural networks are easily fooled: High confidence predictions
  for unrecognizable images.
\newblock In {\em Proceedings of the IEEE Conference on Computer Vision and
  Pattern Recognition}, pages 427--436, June 2015.

\bibitem{oliva2007role}
Aude Oliva and Antonio Torralba.
\newblock The role of context in object recognition.
\newblock {\em Trends in cognitive sciences}, 11(12):520--527, 2007.

\bibitem{papandreou2015weakly}
George Papandreou, Liang-Chieh Chen, Kevin~P Murphy, and Alan~L Yuille.
\newblock Weakly-and semi-supervised learning of a deep convolutional network
  for semantic image segmentation.
\newblock In {\em Proceedings of the IEEE International Conference on Computer
  Vision}, pages 1742--1750, 2015.

\bibitem{patel2015visual}
Vishal~M Patel, Raghuraman Gopalan, Ruonan Li, and Rama Chellappa.
\newblock Visual domain adaptation: A survey of recent advances.
\newblock {\em IEEE signal processing magazine}, 32(3):53--69, 2015.

\bibitem{PengSAS15}
Xingchao Peng, Baochen Sun, Karim Ali, and Kate Saenko.
\newblock Learning deep object detectors from 3d models.
\newblock In {\em 2015 {IEEE} International Conference on Computer Vision,
  {ICCV} 2015, Santiago, Chile, December 7-13, 2015}, pages 1278--1286, 2015.

\bibitem{piaget2013construction}
Jean Piaget.
\newblock {\em The construction of reality in the child}, volume~82.
\newblock Routledge, 2013.

\bibitem{pillai2015monocular}
Sudeep Pillai and John Leonard.
\newblock Monocular slam supported object recognition.
\newblock In {\em Robotics: Science and Systems}, 2015.

\bibitem{nips17-ws-robots}
Ingmar Posner, Raia Hadsell, Martin Riedmiller, Markus Wulfmeier, and Rohan
  Paul.
\newblock {Neural Information Processing Systems (NIPS) Workshop on Acting and
  Interacting in the Real World: Challenges in Robot Learning}.
\newblock \url{http://sites.google.com/view/nips17robotlearning/home}, 2016.

\bibitem{povinelli2000}
Daniel~J Povinelli.
\newblock Folk physics for apes: The chimpanzee's theory of how the world
  works.
\newblock 2000.

\bibitem{rasmus2015semi}
Antti Rasmus, Mathias Berglund, Mikko Honkala, Harri Valpola, and Tapani Raiko.
\newblock Semi-supervised learning with ladder networks.
\newblock In {\em Advances in Neural Information Processing Systems}, pages
  3546--3554, 2015.

\bibitem{Rebuffi_2017_CVPR}
Sylvestre-Alvise Rebuffi, Alexander Kolesnikov, Georg Sperl, and Christoph~H.
  Lampert.
\newblock {}icarl: Incremental classifier and representation learning.

\bibitem{redmon2016yolo9000}
Joseph Redmon and Ali Farhadi.
\newblock Yolo9000: better, faster, stronger.
\newblock In {\em IEEE Conference on Computer Vision and Pattern Recognition
  (CVPR)}, 2016.

\bibitem{rezende2016one}
Danilo Rezende, Ivo Danihelka, Karol Gregor, Daan Wierstra, et~al.
\newblock One-shot generalization in deep generative models.
\newblock In {\em International Conference on Machine Learning}, pages
  1521--1529, 2016.

\bibitem{DBLP:journals/corr/RichardWebsterA16}
Brandon RichardWebster, Samuel~E. Anthony, and Walter~J. Scheirer.
\newblock Psyphy: {A} psychophysics driven evaluation framework for visual
  recognition.
\newblock {\em CoRR}, abs/1611.06448, 2016.

\bibitem{rock1983logic}
I.~Rock.
\newblock {\em The logic of perception}.
\newblock Cambridge: MIT Press, 1983.

\bibitem{rumsfeld-known-knowns}
Donald Rumsfeld.
\newblock {DoD News Briefing} addressing {\em unknown unknowns.}, 2002.
\newblock Accessed 2016-12-21 via \url{htp://goo.gl/ph7UfV}.

\bibitem{rusu2016progressive}
Andrei Rusu, Neil Rabinowitz, Guillaume Desjardins, Hubert Soyer, James
  Kirkpatrick, Koray Kavukcuoglu, Razvan Pascanu, and Raia Hadsell.
\newblock Progressive neural networks.
\newblock {\em arXiv preprint arXiv:1606.04671}, 2016.

\bibitem{sadeghi2016cad2rl}
Fereshteh Sadeghi and Sergey Levine.
\newblock Cad2rl: Real single-image flight without a single real image.
\newblock {\em arXiv preprint arXiv:1611.04201}, 2016.

\bibitem{salas2013slam++}
Renato~F Salas-Moreno, Richard~A Newcombe, Hauke Strasdat, Paul~HJ Kelly, and
  Andrew~J Davison.
\newblock Slam++: Simultaneous localisation and mapping at the level of
  objects.
\newblock In {\em Proceedings of the IEEE conference on computer vision and
  pattern recognition}, pages 1352--1359, 2013.

\bibitem{santoro2016meta}
Adam Santoro, Sergey Bartunov, Matthew Botvinick, Daan Wierstra, and Timothy
  Lillicrap.
\newblock Meta-learning with memory-augmented neural networks.
\newblock In {\em International conference on machine learning}, pages
  1842--1850, 2016.

\bibitem{6701391}
Walter~J. Scheirer, Sam~E. Anthony, Ken Nakayama, and David~D. Cox.
\newblock Perceptual annotation: Measuring human vision to improve computer
  vision.
\newblock {\em IEEE Transactions on Pattern Analysis and Machine Intelligence},
  36(8):1679--1686, Aug 2014.

\bibitem{scheirer2013toward}
Walter~J Scheirer, Anderson de~Rezende~Rocha, Archana Sapkota, and Terrance~E
  Boult.
\newblock Toward open set recognition.
\newblock {\em IEEE Transactions on Pattern Analysis and Machine Intelligence},
  35(7):1757--1772, 2013.

\bibitem{6365193}
Walter~J. Scheirer, Anderson de~Rezende~Rocha, Archana Sapkota, and Terrance~E.
  Boult.
\newblock Toward open set recognition.
\newblock {\em IEEE Transactions on Pattern Analysis and Machine Intelligence},
  35(7):1757--1772, July 2013.

\bibitem{6809169}
Walter~J. Scheirer, Lalit~P. Jain, and Terrance~E. Boult.
\newblock Probability models for open set recognition.
\newblock {\em IEEE Transactions on Pattern Analysis and Machine Intelligence},
  36(11):2317--2324, Nov 2014.

\bibitem{schierwagen_reverse_2012}
Andreas Schierwagen.
\newblock On reverse engineering in the brain and cognitive sciences.
\newblock {\em Natural Computing}, 11(1):141--150, 2012.

\bibitem{Sch15Rob}
T.~Schmidt, K.~Hertkorn, R.~Newcombe, Z.~Marton, S.~Suppa, and D.~Fox.
\newblock {Robust Real-Time Tracking with Visual and Physical Constraints for
  Robot Manipulation}.
\newblock In {\em {Proc.~of the IEEE International Conference on Robotics \&
  Automation (ICRA)}}, 2015.

\bibitem{Schulmanetal_ICML2015}
John Schulman, Sergey Levine, Philipp Moritz, Michael~I. Jordan, and Pieter
  Abbeel.
\newblock Trust region policy optimization.
\newblock In {\em Proceedings of the 32nd International Conference on Machine
  Learning (ICML)}, 2015.

\bibitem{Schulmanetal_ICLR2016}
John Schulman, Philipp Moritz, Sergey Levine, Michael Jordan, and Pieter
  Abbeel.
\newblock High-dimensional continuous control using generalized advantage
  estimation.
\newblock In {\em Proceedings of the International Conference on Learning
  Representations (ICLR)}, 2016.

\bibitem{Shi-2009}
Lei Shi and Thomas~L. Griffiths.
\newblock Neural implementation of hierarchical bayesian inference by
  importance sampling.
\newblock In {\em Proceedings of the Neural Information Processing Systems
  Conference (NIPS)}, 2009.

\bibitem{Simon-96}
Herbert~A. Simon.
\newblock {\em The Sciences of the Artificial}.
\newblock MIT Press, 1996.

\bibitem{SuQLG15}
Hao Su, Charles~Ruizhongtai Qi, Yangyan Li, and Leonidas~J. Guibas.
\newblock Render for {CNN:} viewpoint estimation in images using cnns trained
  with rendered 3d model views.
\newblock In {\em 2015 {IEEE} International Conference on Computer Vision,
  {ICCV} 2015, Santiago, Chile, December 7-13, 2015}, pages 2686--2694, 2015.

\bibitem{suenderhauf2016place}
Niko S{\"u}nderhauf, Feras Dayoub, Sean McMahon, Ben Talbot, Ruth Schulz, Peter
  Corke, Gordon Wyeth, Ben Upcroft, and Michael Milford.
\newblock Place categorization and semantic mapping on a mobile robot.
\newblock In {\em Robotics and Automation (ICRA), 2016 IEEE International
  Conference on}, pages 5729--5736. IEEE, 2016.

\bibitem{rss17-ws-deepLearning}
Niko S\"underhauf, J\"urgen Leitner, Pieter Abbeel, Michael Milford, and Peter
  Corke.
\newblock {Robotics: Science and Systems (RSS) Workshop on New Frontiers for
  Deep Learning in Robotics}.
\newblock \url{http://juxi.net/workshop/deep-learning-rss-2017/}, 2017.

\bibitem{rss16-ws-skeptics}
Niko S\"underhauf, J\"urgen Leitner, Michael Milford, Ben Upcroft, Pieter
  Abbeel, Wolfram Burgard, and Peter Corke.
\newblock {Robotics: Science and Systems (RSS) Workshop Are the Sceptics Right?
  Limits and Potentials of Deep Learning in Robotics}.
\newblock \url{http://juxi.net/workshop/deep-learning-rss-2016/}, 2016.

\bibitem{sunderhauf2016meaningful}
Niko S{\"u}nderhauf, Trung~T Pham, Yasir Latif, Michael Milford, and Ian Reid.
\newblock Meaningful maps - object-oriented semantic mapping.
\newblock In {\em International Conference on Intelligent Robots and Systems
  (IROS)}, 2017.

\bibitem{DBLP:journals/corr/SzegedyZSBEGF13}
Christian Szegedy, Wojciech Zaremba, Ilya Sutskever, Joan Bruna, Dumitru Erhan,
  Ian~J. Goodfellow, and Rob Fergus.
\newblock Intriguing properties of neural networks.
\newblock {\em CoRR}, abs/1312.6199, 2013.

\bibitem{Tamar-2016}
Aviv Tamar, Sergey Levine, and Pieter Abbeel.
\newblock Value iteration networks.
\newblock {\em CoRR}, abs/1602.02867, 2016.

\bibitem{Thrun05}
Sebastian Thrun, Wolfram Burgard, and Dieter Fox.
\newblock {\em Probabilistic Robotics}.
\newblock The MIT Press, 2005.

\bibitem{tobin2017domain}
Josh Tobin, Rachel Fong, Alex Ray, Jonas Schneider, Wojciech Zaremba, and
  Pieter Abbeel.
\newblock Domain randomization for transferring deep neural networks from
  simulation to the real world.
\newblock In {\em Intelligent Robots and Systems (IROS), 2017 IEEE/RSJ
  International Conference on}, pages 23--30. IEEE, 2017.

\bibitem{todorov2012}
E.~Todorov, T.~Erez, and Y.~Tassa.
\newblock Mujoco: A physics engine for model-based control.
\newblock In {\em International Conference on Intelligent Robots and Systems
  {IROS}}, 2012.

\bibitem{torralba2011unbiased}
Antonio Torralba and Alexei~A Efros.
\newblock Unbiased look at dataset bias.
\newblock In {\em Computer Vision and Pattern Recognition (CVPR)}, pages
  1521--1528. IEEE, 2011.

\bibitem{TzengDHFPLSD15}
Eric Tzeng, Coline Devin, Judy Hoffman, Chelsea Finn, Xingchao Peng, Sergey
  Levine, Kate Saenko, and Trevor Darrell.
\newblock Towards adapting deep visuomotor representations from simulated to
  real environments.
\newblock {\em CoRR}, abs/1511.07111, 2015.

\bibitem{TzengHDS15}
Eric Tzeng, Judy Hoffman, Trevor Darrell, and Kate Saenko.
\newblock Simultaneous deep transfer across domains and tasks.
\newblock In {\em 2015 {IEEE} International Conference on Computer Vision,
  {ICCV} 2015, Santiago, Chile, December 7-13, 2015}, pages 4068--4076, 2015.

\bibitem{TzengHZSD14}
Eric Tzeng, Judy Hoffman, Ning Zhang, Kate Saenko, and Trevor Darrell.
\newblock Deep domain confusion: Maximizing for domain invariance.
\newblock {\em CoRR}, abs/1412.3474, 2014.

\bibitem{vinyals2016matching}
Oriol Vinyals, Charles Blundell, Tim Lillicrap, Daan Wierstra, et~al.
\newblock Matching networks for one shot learning.
\newblock In {\em Advances in Neural Information Processing Systems}, pages
  3630--3638, 2016.

\bibitem{helmholtz1867handbuch}
Hermann Von~Helmholtz.
\newblock {\em Handbuch der physiologischen Optik}, volume~9.
\newblock Voss, 1867.

\bibitem{wang2016learning}
Yu-Xiong Wang and Martial Hebert.
\newblock Learning to learn: Model regression networks for easy small sample
  learning.
\newblock In {\em European Conference on Computer Vision}, pages 616--634.
  Springer, 2016.

\bibitem{watter2015embed}
Manuel Watter, Jost Springenberg, Joschka Boedecker, and Martin Riedmiller.
\newblock Embed to control: A locally linear latent dynamics model for control
  from raw images.
\newblock In {\em Advances in Neural Information Processing Systems}, pages
  2728--2736, 2015.

\bibitem{Wilson-2009}
Robert~C. Wilson and Leif~H. Finkel.
\newblock A neural implementation of the {Kalman Filter}.
\newblock In {\em Proceedings of the Neural Information Processing Systems
  Conference (NIPS)}, 2009.

\bibitem{wu2015galileo}
Jiajun Wu, Ilker Yildirim, Joseph~J Lim, Bill Freeman, and Josh Tenenbaum.
\newblock Galileo: Perceiving physical object properties by integrating a
  physics engine with deep learning.
\newblock In {\em Advances in Neural Information Processing Systems}, pages
  127--135, 2015.

\bibitem{Wu_2015_CVPR}
Zhirong Wu, Shuran Song, Aditya Khosla, Fisher Yu, Linguang Zhang, Xiaoou Tang,
  and Jianxiong Xiao.
\newblock 3d shapenets: A deep representation for volumetric shapes.
\newblock In {\em The IEEE Conference on Computer Vision and Pattern
  Recognition (CVPR)}, June 2015.

\bibitem{yan2016perspective}
Xinchen Yan, Jimei Yang, Ersin Yumer, Yijie Guo, and Honglak Lee.
\newblock {Perspective transformer nets: Learning single-view 3d object
  reconstruction without 3d supervision}.
\newblock In {\em Advances in Neural Information Processing Systems (NIPS)},
  pages 1696--1704, 2016.

\bibitem{yildirim2017physical}
Ilker Yildirim, Tobias Gerstenberg, Basil Saeed, Marc Toussaint, and Josh
  Tenenbaum.
\newblock Physical problem solving: Joint planning with symbolic, geometric,
  and dynamic constraints.
\newblock {\em arXiv preprint arXiv:1707.08212}, 2017.

\bibitem{Zhang-2016}
Chiyuan Zhang, Samy Bengio, Moritz Hardt, Benjamin Recht, and Oriol Vinyals.
\newblock Understanding deep learning requires rethinking generalization.
\newblock {\em CoRR}, abs/1611.03530, 2016.

\bibitem{zhang2015towards}
Fangyi Zhang, J{\"u}rgen Leitner, Michael Milford, Ben Upcroft, and Peter
  Corke.
\newblock Towards vision-based deep reinforcement learning for robotic motion
  control.
\newblock In {\em Australasian Conference on Robotics and Automation}, 2015.

\bibitem{7577876}
He~Zhang and Vishal Patel.
\newblock Sparse representation-based open set recognition.
\newblock {\em IEEE Transactions on Pattern Analysis and Machine Intelligence},
  PP(99):1--1, 2016.

\bibitem{zhang2016deepcontext}
Yinda Zhang, Mingru Bai, Pushmeet Kohli, Shahram Izadi, and Jianxiong Xiao.
\newblock Deepcontext: context-encoding neural pathways for 3d holistic scene
  understanding.
\newblock In {\em IEEE International Conference on Computer Vision (ICCV)},
  2017.

\bibitem{zhu2017rethinking}
Rui Zhu, Hamed~Kiani Galoogahi, Chaoyang Wang, and Simon Lucey.
\newblock Rethinking reprojection: Closing the loop for pose-aware shape
  reconstruction from a single image.
\newblock In {\em IEEE International Conference on Computer Vision (ICCV)},
  pages 57--65. IEEE, 2017.

\bibitem{ZhuMKLGFF16}
Yuke Zhu, Roozbeh Mottaghi, Eric Kolve, Joseph~J. Lim, Abhinav Gupta,
  Li~Fei{-}Fei, and Ali Farhadi.
\newblock Target-driven visual navigation in indoor scenes using deep
  reinforcement learning.
\newblock {\em CoRR}, abs/1609.05143, 2016.

\end{thebibliography}

\end{document}